# Kernel principal component analysis network for image classification[1]

Wu Dan[1,4]   Wu Jiasong[1,2,3,4]   Zeng Rui[1,4]   Jiang Longyu[1,4]   Lotfi Senhadji[2,3,4]   Shu Huazhong[1,4]

([1] Key Laboratory of Computer Network and Information Integration of Ministry of Education, Southeast University, Nanjing 210096, China)

([2] Institut National de la Santé et de la Recherche Médicale U 1099, Rennes 35000, France)

([3] Laboratoire Traitement du Signal et de l'Image, Université de Rennes 1, Rennes 35000, France)

([4] Centre de Recherche en Information Biomédicale Sino-français, Nanjing 210096, China)

**Abstract**: In order to classify the nonlinear feature with linear classifier and improve the classification accuracy, a deep learning network named kernel principal component analysis network (KPCANet) is proposed. First, mapping the data into higher space with kernel principal component analysis to make the data linearly separable. Then building a two-layer KPCANet to obtain the principal components of image. Finally, classifying the principal components with linearly classifier. Experimental results show that the proposed KPCANet is effective in face recognition, object recognition and hand-writing digits recognition, it also outperforms principal component analysis network (PCANet) generally as well. Besides, KPCANet is invariant to illumination and stable to occlusion and slight deformation。

A major difficulty of image classification comes from the considerable intra-class variability, arising from different illuminations, rigid deformations, non-rigid deformations and occlusions, which are useless for classification and should be eliminated. Deep learning structures like deep convolutional networks have the ability to learn invariant features [1]. Bruna et al [2] built a scattering network (ScatNet) that is invariant to both rigid and also non-rigid deformations. Chan et al.[3] constructed a principal component analysis network (PCANet), which cascaded principal component analysis (PCA), binary hashing, and block-wise histogram. PCANet achieves the state of the art accuracy in many datasets of classification tasks, like extended Yale B dataset, AR dataset, and FERET dataset. Kernel PCA (KPCA)[4-5] is a nonlinear generalization of PCA in the sense that it performs PCA in the feature spaces of arbitrary large dimension. KPCA can generally provide better recognition rate than the ordinary PCA by the following two reasons: 1) KPCA uses arbitrary number of nonlinear components, however, ordinary PCA uses only limited number of linear principal components; 2) KPCA has more flexibility than ordinary PCA since KPCA can choose different kernel functions (for example, Gaussian kernel, Polynomial kernel, etc.) for different recognition tasks, however, ordinary PCA uses only the linear kernel.

In this paper, we propose a new deep learning network named kernel principal component network (KPCANet), which cascades two KPCA stages and one pooling stage. When the kernel function is linear, the proposed KPCANet degrades to the PCANet[3]. Experimental results show that the proposed KPCANet is invariant to illumination and stable to slight non-rigid deformation and generally outperforms PCANet in both face recognition and object recognition tasks.

## 1 KPCANet

Fig.1 shows the whole structure of the proposed KPCANet, which consists of two KPCA stages and one pooling stage. Suppose that the patch size is $k_1 \times k_2$ at all stages, and all the input images are of size $m \times n$.

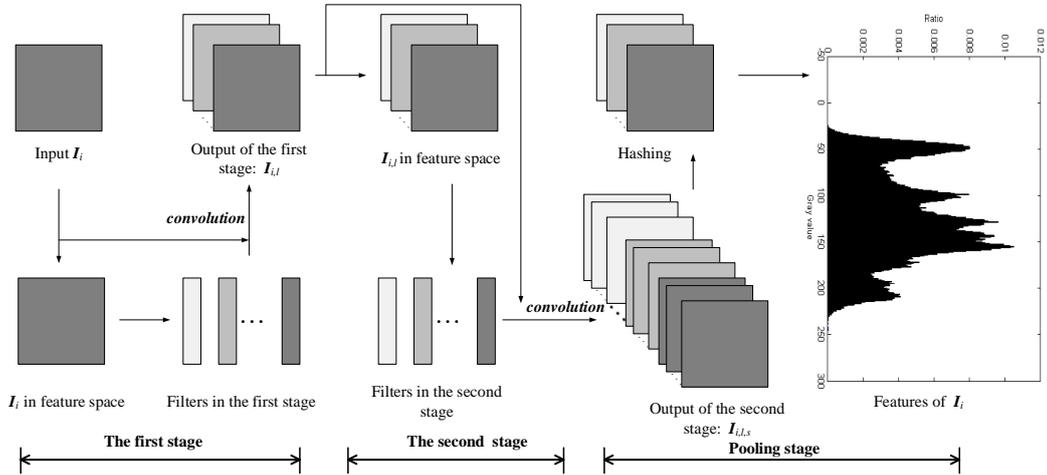

**Fig.1** The detailed block diagram of the proposed KPCANet

### 1.1 The first stage of KPCANet

Inputting $N$ images $I_i (i=1,2,\ldots,N)$ that belong to $c$ classes. We take a patch $p_{i,j} \in \mathbf{R}^{k_1 \times k_2}$ centered in the $j$-th ($j=1,2,\ldots,mn$) pixel of image $I_i$ and vectorize the patch as $x_{i,j} \in \mathbf{R}^{k_1 k_2}$. Collecting all the vectorized patches $x_{i,j}, j=1,2,\ldots,mn$ of $I_i$, we obtain a matrix $X_i = [x_{i,1}, x_{i,2}, \ldots, x_{i,mn}] \in \mathbf{R}^{k_1 k_2 \times mn}$. Construct the same matrix for all input images and put them together, we get $X = [X_1, X_2, \ldots, X_N] \in \mathbf{R}^{k_1 k_2 \times Nmn}$. For convenience, the $p$-th column of $X$ is denoted as $x'_p, p = 1, 2, \ldots, Nmn$. We then map $X$ from the input space $\mathbb{R}^{k_1 k_2 \times k_1 k_2}$ to the feature space $\mathcal{F}$ by

$$T : \mathbb{R}^{k_1 k_2 \times k_1 k_2} \to \mathcal{F}, X \mapsto X^{\mathcal{F}} \tag{1}$$

To find the principal component of $T(x'_p)$, we need to diagonalize the covariance matrix $C$:

$$C = \frac{1}{Nmn} \sum_{p=1}^{Nmn} T(x'_p) T(x'_p)^T \tag{2}$$

To simplify the diagonalization of C, we could diagonalize $K$ instead, where $K_{pq} = (\tilde{T}(x'_p) \cdot \tilde{T}(x'_q))_{pq}$, $\tilde{T}(x'_p)$ denotes the centralised $T(x'_p)$ and symbol "$\cdot$" denotes the dot product. Since the dimension of $\mathcal{F}$ could be arbitrarily large even infinite [4-5], it would be difficult to compute dot product $(\tilde{T}(x'_p) \cdot \tilde{T}(x'_q))$ directly, therefore, we substitute dot product with kernel function $k$ and obtain $K_{pq} = (k(x'_p, x'_q))_{pq}$. After that, $K$ is centralized with $K' = K - 1_{Nmn} K - K 1_{Nmn} + 1_{Nmn} K 1_{Nmn}$ and $K'$ is diagonalized to get the principal eigenvectors $W_l^1, l = 1, 2, \ldots, L_1$, which is the KPCA filters in the first stage as well, where $(1_{Nmn})_{ij} = \frac{1}{Nmn}$.

Zero-padding the boundary of $I_i$ and convolve it with $W_l^1$, we get the $l$-th filter output of the first stage $I_{i,l} = I_i * W_l^1 \in \mathbf{R}^{m \times n}, i = 1, 2, \ldots, N; l = 1, 2, \ldots, L_1$, where "$*$" denotes 2D convolution and $L_1$ denotes the amount of filters in the first stage..

### 1.2 The second stage of KPCANet

By repeating the same process as in the first stage on $I_{i,l}, i = 1, 2, \ldots, N; l = 1, 2, \ldots, L_1$, we obtain $L_2$ kernel PCA filters $W_s^2, s = 1, 2, \ldots, L_2$ of the second stage. Convolve $I_{i,l}$ with $W_s^2$, we get the output of the second stage $I_{i,l,s} = I_{i,l} * W_s^2, i = 1, 2, \ldots, N; l = 1, 2, \ldots, L_1; s = 1, 2, \ldots, L_2$;

### 1.3 The pooling stage of KPCANet

Every $L_2$ input image is binarized and converted to an image with:

$$\begin{aligned} P_{i,l} &= \sum_{s=1}^{L_2} 2^{s-1} H(I_{i,l,s}) \\ &= \sum_{s=1}^{L_2} 2^{s-1} H(I_{i,l,s}) I_{i,l} * W_s^2, i = 1, 2, \ldots, N; l = 1, 2, \ldots, L_1; \end{aligned} \tag{3}$$

where $H$ is the Heaviside step (like) function [3].

Each of the $L_1$ images $P_{i,l}(l = 1, 2, \ldots, L_1)$ is then partitioned into $B$ blocks. We compute the histogram of the decimal values in each block, and concatenate all the $B$ histograms into one vector denoted as $Bhist(P_{i,l})$. At last, the KPCANet features of $I_i$ are given by

$$f_i = \left[ Bhist(P_{i,1}), Bhist(P_{i,2}), \ldots, Bhist(P_{i,L_1}) \right]^T \in \mathbf{R}^{(2^{L_2})L_1 B} \qquad (4)$$

Since deep architectures are composed of multiple levels of nonlinear operations, such as in complicated propositional formulae re-using many sub-formulae[6], the first two stages of KPCANet are set to be the same in this paper, we could re-use the whole structure of the first stage as well.

Tab. 1  Various kernel functions used in this paper

| Kernel function | Expression | Value of parameters |
|---|---|---|
| Linear | $k(x,y) = xy + c$ | $c = 0$ |
| Gaussian | $k(x,y) = exp\left(-\dfrac{\|x-y\|^2}{2\sigma^2}\right)$ | $\sigma = 1$ |
| Polynomial | $k(x,y) = (xy + 1)^d$ | $d = 3$ |
| Exponential | $k(x,y) = exp\left(-\dfrac{\|x-y\|}{2\sigma^2}\right)$ | $\sigma = 1$ |
| Laplacian | $k(x,y) = exp\left(-\dfrac{\|x-y\|}{\sigma}\right)$ | $\sigma = 1$ |
| Sigmoid | $k(x,y) = \tanh(\alpha xy + c)$ | $\alpha = \dfrac{1}{2}, c = -1$ |
| Rational quadratic | $k(x,y) = exp\left(1 - \dfrac{\|x-y\|^2}{\|x-y\|^2 + c}\right)$ | $c = 1$ |
| Inverse multiquadric | $k(x,y) = \dfrac{1}{\sqrt{\|x-y\|^2 + c^2}}$ | $c = 1$ |
| Circular | $k(x,y) = \begin{cases} \dfrac{2}{\pi}\text{acos}\left(\dfrac{\|x-y\|}{\sigma}\right) - \dfrac{2}{\pi}\|x-y\|\sqrt{\dfrac{\|x-y\|^2}{\sigma^2}}, & \|x-y\| < \sigma \\ 0, & otherwise \end{cases}$ | $\sigma = 0.2$ |
| Spherical | $k(x,y) = \begin{cases} 1 - \dfrac{3}{2}\dfrac{\|x-y\|}{\sigma} + \dfrac{1}{2}\left(\dfrac{\|x-y\|}{\sigma}\right)^3, & \|x-y\| < \sigma \\ 0, & otherwise \end{cases}$ | $\sigma = 0.2$ |

## 2  Experimental Results

We now evaluate the performance of the proposed KPCANet on various databases including MNIST, USPS, Yale face dataset, COIL-100 objects dataset, and AR dataset. Besides, we compared KPCANets that cascade various (from one to three) stage(s) of KPCA layer in this paper as well. All the features learned by KPCANet are classified with SVM classifier.

In this section, we use various kernel functions to evaluate the performance of the proposed KPCANet in recognition tasks including hand-written digits recognition, face recognition and object recognition. Kernel functions that are used in this paper are presented in Tab. 1.

MNIST[7] and USPS[8] are used to evaluate the performance of KPCANet on hand-written images. MNIST contains 60000 train images and 10000 test images, all images are of size 28×28. USPS contains 9298 images of size 16×16 in total, 5000 of them are chosen randomly to train KPCANet, and the rest are for testing. The Yale Face Database[9] is used to evaluate the performance of proposed KPCANet on face images, it contains 165 grayscale images in GIF format of 15 individuals, each individual contains 11 images, with different facial expression or configuration: center-light, wear glasses, happy, left-light, wear no glasses, normal, right-light, sad, sleepy, surprised, and wink. All images of this database are cropped to size 64×64, and 90 of them are chosen randomly to train the proposed KPCANet, the rest are for testing. COIL-100 (Columbia Object Image Library)[10] is a database of color images of 100 objects. Images of the objects are taken at pose intervals of 5 degrees, this corresponds to 72 poses per object. All images are transformed into gray images and cropped to size 32×32. Half images of each object are chosen randomly to train KPCANet and the others are for testing.

The performances of different kernel functions on datasets including MNIST, USPS, Yale face dataset and COIL-100 dataset are presented in Tab.2. Both the patch size and the block size are set to 8×8, and the filter numbers are set to 8 at all stages, the overlapping ratio of block is 0.5.

It can be seen from Tab.2 that the performance of PCANet performs better than KPCANets in hand-written digit recognition generally, while the latter outperforms the former in face recognition and object recognition.

Tab. 2 Comparison of error rates of KPCANet with various kernel functions on different datasets    %

| Kernel function | MNIST | USPS | Yale face dataset | COIL-100 objects dataset |
| --- | --- | --- | --- | --- |
| Linear | 0.85 | 2.37 | 5.33 | 1.36 |
| Gaussian | 1.07 | 2.70 | 4.00 | 0.89 |
| Polynomial | 1.06 | 2.68 | 2.67 | 1.50 |
| Exponential | 0.97 | 2.51 | 8.00 | 1.19 |
| Laplacian | 1.06 | 2.84 | 4.00 | 1.14 |
| Rational quadratic | 1.08 | 2.61 | 4.00 | 1.42 |
| Sigmoid | 0.98 | 2.40 | 2.67 | 1.53 |
| Inverse multiquadric | 0.77 | 2.61 | 4.00 | 1.61 |
| Circular | 1.03 | 2.58 | 4.00 | 1.64 |
| Spherical | 0.88 | 2.56 | 5.33 | 1.50 |

**2.1 Face recognition on AR face dataset**

The properties of KPCANet that invariant to illumination and stable to slight deformations and occlusions are tested by performing KPCANet on AR dataset[11] in this section.

AR dataset contains about 4000 color images from 126 individuals. The subset of the data that contains 100 individuals consisting of 50 males and 50 females of size 165×120 is chosen. The color images are converted to gray scale ones. Each individual consists of 2 images with frontal illumination and neutral expression, which is used as the training sample, the other images, including 24 images variation from illumination to disguise, are used for testing.

The patch size and the block size are set to be 7×7 and 8×8, respectively. The overlapping ratio of block is 0.5. We compare the proposed KPCANet with LBP[12] and P-LBP[13] in Tab.3. KPCANet with linear kernel function and Laplacian kernel function is used in this experiment. From Tab.3, one can see that when the images only undergo the change of illumination, the testing accuracy rate achieves 100% with both linear kernel KPCANet and Laplacian kernel KPCANet, this demonstrates that KPCANet is invariant to illumination. Besides, KPCANet outperforms LBP[12] and P-LBP[13] on different expressions and disguises with various illumination conditions, which show that KPCANet is robust to small deformation and occlusion.

Tab. 3 Comparison of accuracy rates of the methods on AR face database    %

| Test sets | Illumination | Expression | Disguise with illumination |
|---|---|---|---|
| LBP [12] | 93.83 | 81.33 | 83.50 |
| P-LBP[13] | 97.50 | 80.33 | 90.05 |
| KPCANet (linear kernel) | 100.00 | 95.67 | 99.50 |
| KPCANet (Laplacian kernel) | 100.00 | 94.33 | 99.59 |

### 2.2 KPCANet with various stages in AR face dataset

In this section, we perform KPCANet which cascade different number of KPCA filter bank layer and a pooling layer with AR face dataset we used in Section 2.1, and all images are cropped to size 32×32. Linear kernel, sigmoid kernel and circular kernel are chosen here in order to simplify the result. The patch size and the block size are set to be 7×7 and 8×6, respectively. The overlapping ratio of block is 0.5. The results are shown in Tab.4.

Tab.4 Comparison of accuracy rates of KPCANet with different stages number on AR face dataset    %

| Filter bank layer number | Linear kernel | Sigmoid kernel | Circular kernel |
|---|---|---|---|
| One | 94.67 | 94.67 | 92.00 |
| Two | 96.00 | 94.67 | 94.67 |
| three | 96.00 | 97.33 | 96.00 |

From Tab.4, we can see that the accuracy rate increase as the number of KPCA filter bank layer increase in KPCANet, however, training time grows exponentially at the same time.

### 3  Conclusion

In this paper, we propose the KPCANet, which is an extension of PCANet, for image classification. The proposed KPCANet cascades kernel PCA, binary hashing and block-wise histogram. Experiments prove that KPCANet with different kernel functions is stable in general and also is invariant to illumination and stable to slight deformation and occlusion. Moreover, KPCANet is suitable for the recognition of hand-written images, face images and object images.